\documentclass[citeauthoryear]{llncs}
\usepackage{graphicx}
\usepackage{multirow}
\usepackage{booktabs}
\usepackage{setspace}

\begin{document}
\title{Maize Haploid Identification via LSTM-CNN and Hyperspectral Imaging Technology}
\author{Xuan-Yu Wang\orcidID{0000-0002-8753-6742} \and
	Wen-Xuan Liao\orcidID{0000-0003-4899-5468} \and { } Dong An\orcidID{0000-0002-8389-987X}\inst{\star} \and
	Yao-Guang Wei\orcidID{0000-0003-4888-8558}\thanks{Corresponding author}}

\authorrunning{Wang et al.}

\institute{College of Information and Electriacl Engineering, China Agricultural University, Beijing 100083, China \email{\{wangxuanyu,vane,andong,weiyaoguang\}@cau.edu.cn}} 

\maketitle

\begin{abstract}
Accurate and fast identification of seed cultivars is crucial to plant breeding, with accelerating breeding of new products and increasing its quality. In our study, the first attempt to design a high-accurate identification model of maize haploid seeds from diploid ones based on optimum waveband selection of the LSTM-CNN algorithm is realized via deep learning and hyperspectral imaging technology, with accuracy reaching 97\% in the determining optimum waveband of 1367.6-1526.4nm. The verification of testing another cultivar achieved an accuracy of 93\% in the same waveband. The model collected images of 256 wavebands of seeds in the spectral region of 862.9-1704.2nm. The high-noise waveband intervals were found and deleted by the LSTM. The optimum-data waveband intervals were determined by CNN's waveband-based detection. The optimum sample set for network training only accounted for 1/5 of total sample data. The accuracy was significantly higher than the full-waveband modeling or modeling of any other wavebands. Our study demonstrates that the proposed model has outstanding effect on maize haploid identification and it could be generalized to some extent.

\end{abstract}
\section{Introduction}
The seed is the basis of agricultural engineering. More efficient breeding strategies of new cultivars, especially the identification and accuracy of the target cultivar, are desired due to the increasing population and changing climates. Because mixing cultivars decreases the purity of breeding experiment and crop output. The maize haploid breeding technology, which can shorten breeding time and improve germplasm, has become the key in breeding new maize (Murigneux et al. 1993). However, the occurrence rate of maize haploid under natural conditions is about 1\%, and it can increase to 8\%-15\% after artificial induction (Chalyk et al. 2001; Chen et al. 2003; Prigge et al. 2011). Therefore, how to identify maize haploid seeds fast, accurately and noninvasively is significant. \\
Seeds are traditionally classified and identified through morphology method (Sanchez et al. 1993), protein electrophoresis (Arun et al. 2010), DNA molecular marker technology tests (Ye et al. 2013), genetic marker method (Wang et al. 2016) and oil content marker method (Melchinger et al. 2013). However, the first three methods are expensive and time-consuming and require practiced operators (Mahesh et al. 2015). The genetic marker method requires visual identification of difference of genetically expressed characters, which neither achieves accurate identification because of visual fatigue nor suits automatic machine vision sorting (Melchinger et al. 2013). The oil content marker methods identifies haploids through the xenia effect of oil content (Chen et al. 2003). However, there is certain fluctuation on macrostatistics of oil content. Models will make error identification of haploid and diploid due to overlapping of oil contents sometimes (Cui et al. 2017). The noninvasively information acquisition technology mainly depends on noninvasive optics, for example machine vision technology (Olesen et al. 2011). This technology has limited abilities in information acquisition, since the machine vision system mainly collects external feature information, which makes the information collection less effective when compared with collection by the near infrared spectroscopy technology focusing on seeds’ spectral features related with chemical composition combining the near infrared spectral information with near infrared image information (Wu et al. 2013). The near infrared spectra are very sensitive to alkyl, hydroxyl and amidogens in organics and reflect information of protein, starch, water and fat in samples (Mahesh et al. 2008; Pan et al. 2015). In recent years, the hyperspectral technology has been applied successfully to inspecting food (Gowen et al. 2007), detecting and sorting seeds (Kong et al.2013; Huang et al. 2016). Based on above, the hyperspectral imaging technology is feasible to collect the information on the diversity of haploid and diploid maize seeds’ organics or spatial shapes.\\
Based on existing studies, the classical patter recognition algorithm for identification of hyperspectral seed images consists of ``data preprocess, feature extraction and identification'' modules (Gowen et al. 2007; Emamgholizadeh et al. 2015; Jeong et al. 2015; Cen et al. 2014; Rivero et al. 2012). For instance, Zhang et al. (2012) applied the PCA-GLCM-LS-SVM combined models in seed classification. Besides, the classical pattern recognition has to design complicated decision function and feature selection algorithm, as a model learns nonlinear high-dimensional space through the linear kernel functions based on probability distribution (Boser et al. 1992; Cortes et al. 1995; Bishop 2006). Images collected by hyperspectral imaging technology are high-dimensional tensor with distinct noises (Liu et al. 2012). As a result, existing seed identification models have the poor migration effect among different models and complicated structure. Nonlinear mapping realizes the transformation between the linear function of classification algorithm and the nonlinear high-dimensional information space. Traditional classical pattern recognition chooses manual design of nonlinear mapping. Hence, the designed model is only applicable to a specific field. In contrast, the deep learning choose the universal nonlinear mapping which is set hidden on the kernel machine in the layers, with adequate dimension able to cover and fit various training set (Goodfellow et al. 2016). Deep learning training applies the iteration optimization based on gradient. It only has to assure that the cost function converges to a minimum value rather than the global convergence like training logic regression or linear regression (Goodfellow et al. 2016). These structural advantages make the strong generalization and simple structure of the deep learning. Based on above analysis, the deep learning were applied to establish the maize haploid seed identification model.\\
This paper is due to design a maize haploid seed identification model via deep learning algorithm and hyperspectral images. To sum up, the major contributions are three-fold:\\
{\renewcommand\baselinestretch{0.5}\selectfont
\begin{itemize}
	\item[$\bullet$]It is the first attempt to design a maize haploid seed identification model applying optimum hyperspectal images data selected by the LSTM-CNN algorithm. The identification accuracy of the model reaches 97\% and the optimum waveband interval of sample information is 1367.6-1526.4nm. 
	\item[$\bullet$]Instead of using all hyperspectral data as the sample set in traditional methods, the proposed model achieves satisfying identification accuracy only with sample data in the optimum waveband interval. It can lower economic costs of hyperspectral images sampling.
	\item[$\bullet$]The proposed LSTM-CNN model provides some references for construction of the identification model based on hyperspectral imaging technology in other fields. It has certain generalization.
\end{itemize}
\par}
\section{Material and Method}
\subsection{Maize seeds}
\begin{figure}
	\includegraphics[width=\textwidth]{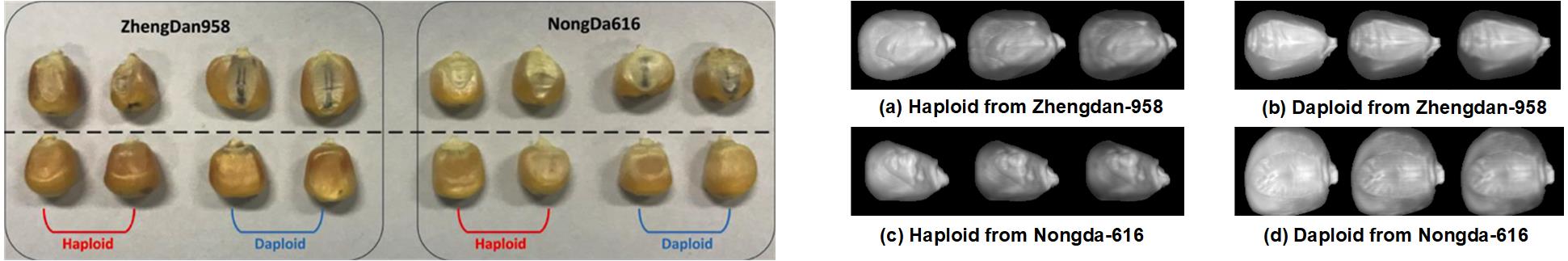}
	\centering
	\caption{Representative samples of haploid and diploid maize seeds and hyperspectral images of 2 cultivars, Zhengdan-958 and Nongda-616. (Left. shows the maize seeds. (Right. shows the haploid or diploid seeds' hyperspectral images taken at 962.7nm, 1132.3nm and 1364.4nm wavebands.} \label{fig1}
\end{figure}

Two maize cultivars (Zhengdan-958 and Nongda-616, Fig.~\ref{fig1}) produced by the National Maize Improvement Center of China Agricultural University from high-oil hybrid induction with R1-nj genetic markers were used as the experimental samples. Cultivars Zhengdan-958 provided 100 haploids and 100 diploids as the experimental samples, so did Nongda-616. Each maize seed was dried, dehydrated, numbered and then stored under 5$^{\circ}$C. Sample images were collected by hyperspectral imaging technology. Considering the great gap between the embryo surface and non-embryo surface, only images of the embryo surface of each seed were used in the experiment. To reduce effects of instrument parameter drifts on measurement results, alternating sampling between haploid and diploid was adopted in actual sampling process. Information of the samples is shown in  Table~\ref{tab1}.\\
\begin{table}
	\centering
	\caption{The number, cultivar and measurment of sample.}\label{tab1}
	\begin{tabular}{c c c c}
		\toprule
		\multicolumn{2}{c}{\bfseries The information of samples} & \multicolumn{2}{c}{\bfseries The number of samples measured}\\
		\midrule
		{\bfseries Acquisition mode} & {\bfseries Cultivar} & {\bfseries Haploid} & {\bfseries Diploid} \\
		\hline
		\multirow{2}{*}{HIS} & Zhengdan-958 & 100 & 100 \\
		\cline{2-4}
		& Nongda-616 & 100 & 100 \\
		\bottomrule
	\end{tabular}
\end{table}
	
\subsection{Hyperspectral imaging technology}
The hyperspectral images were collected by the push-broom GaiaSorter hyperspectral system, which is mainly composed of uniform light source, spectral camera, mobile control platform and the computer. The uniform light source consists of two sets of bromine-tungsten lamps and emits uniform lights through the thermal radiation. The spectral camera is the Image-$\lambda$-N17E “spectrum” near infrared enhanced hyperspectral camera (ZolixInstruments Co., Ltd.). It is integrated with the Imspector series imaging spectrometer and the InGaAs CCD camera. The spectral range of the camera is 862.9-1704.2nm (including 256 wavebands), which covers the near infrared waveband. The spectral resolution, pixel and slit width are 5nm, 320*256 and 30$\mu$m, respectively. The stepping motor controls the system mobile controlling platform and the image acquisition software Spectra View collects images. Under the premise of no image distortion, the moving speed and exposure time of the platform were set 0.27cm/s and 35ms, respectively. All collected images were three-dimensional ones ($x$, $y$, $\lambda$) and the collected image was a (320$\times$2000$\times$256) image cube. To reduce interferences from the external environment, images were collected in a dark box. Measurement errors of the hyperspectral images caused by fluctuation of light source and dark current were corrected by reflection reference on the black and white board according to the Eq.(1):
{\renewcommand\baselinestretch{0.5}\selectfont
\begin{equation}
R_{cur} = \frac{{R_{sam}}-{R_{dar}}}{{R_{whi}}-{R_{dar}}}
\end{equation}
\par}
\noindent where $R_{cur}$ is the calibrated sample image, $R_{sam}$ is the original sample, $R_{dar}$ is the dark reference image and $R_{whi}$ is the white reference image. The dark reference can be acquired when the camera is covered with lens cap. The white reference image can be acquired by one frame that covers the camera completely after the pure white board replacing and lighting the position of the test objects. All sample images after the calibration were used for follow-up experimental analysis.

\subsection{Image segmentation and characteristics extration}
Calibrated sample images still contain the background information. To separate real information of seeds from the background, self-adaptive threshold segmentation and masking (Huang et al. 2016) was used to extract region of interests (ROIs) in following steps: (a) the region of interests (ROIs) extracted in the waveband of 60(1064.8nm) has the highest contrast ratio with the background. The maximum value of background was chosen as the threshold for image binaryzation. (b) The boundary coordinates of each sample were collected, thus getting the binary images. In the same time, the rectangular regions of each sample were determined. Binary masks were generated through the rectangle region to gain ROIs of 256 wavebands. (c) The real information of seeds is the product of multiplying ROIs of each seed and the corresponding binary images, with interferences of background information deleted.
\subsection{High-noise bands removing based on LSTM}
\begin{figure}
	\includegraphics[width=10cm]{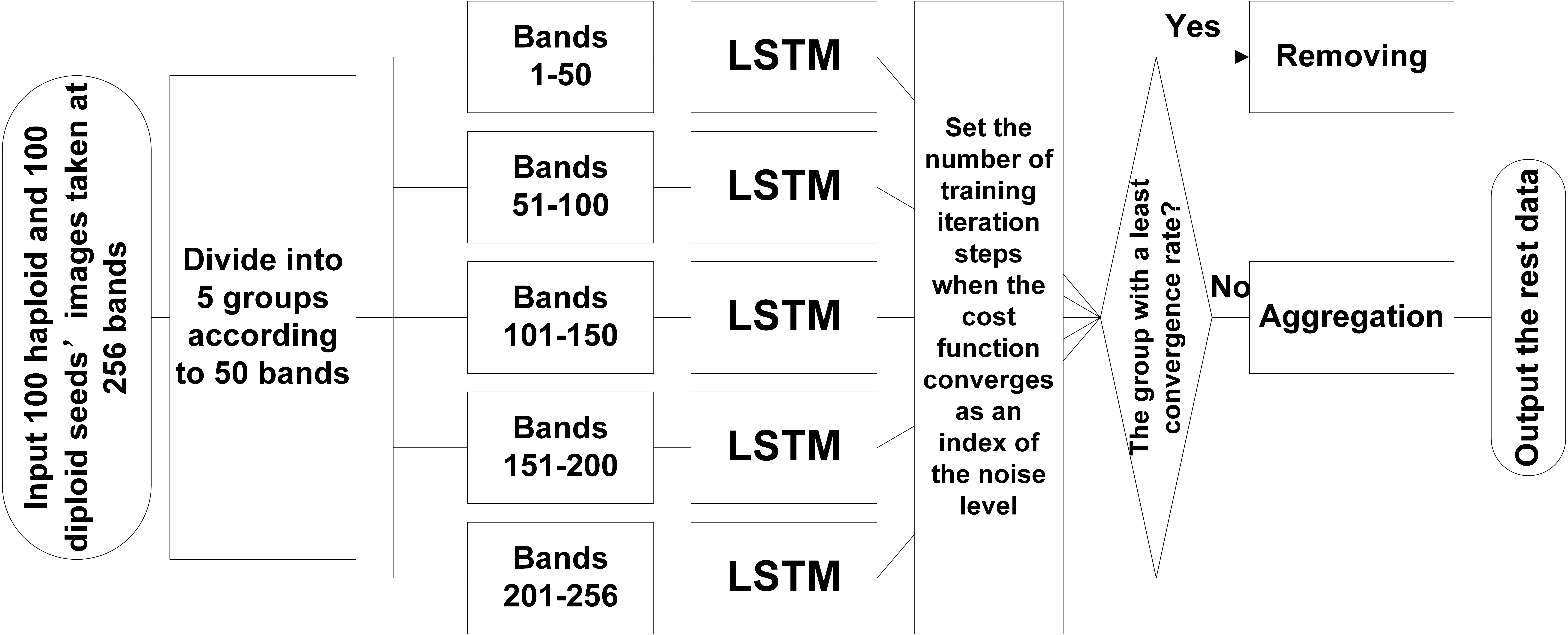}
	\centering
	\caption{Main steps of removing bands including obvious noise. The iterations at convergence of LSTM's cost function is set as the criteria to measure the degree of noises included by the images in each band interval.} \label{fig2}
\end{figure}

Since the imaging spectrometer takes a long time to collect the complete scanning images, electromagnetic radiation in the external space causes complicated impacts on the hyperspectral imaging pathway and brings abundant noise interferences to sample images of some wavebands (Xu et al. 2013). These high noises will cause negative impacts on the training sample images with convolutional neural network (CNN). For example, the final accuracy on the test set is only 0.80 during training of all sample images by the CNN without removing samples including the high-noise. Therefore, it is necessary to design a neural network determining the waveband intervals with high noises and removing them to relieve noise's interferences on the training effect. The CNN fails to achieve satisfying identification accuracy, as the complicated data bring vanishing gradient to the deep CNN during training network with all data. To address this problem, the long short-term memory (LSTM) was applied in this structure. As a RNN network, LSTM can implement single operations on all time step and sequence length and optimize the vanishing gradient with its gating characteristics (Gers et al. 2002). Cell state is the core idea of LSTM. It realizes functions by deleting or adding intracellular states through the input-gate ($i_t$), forget-gate ($f_t$), output-gate ($o_t$) during time accumulating. 
\noindent Specific equations are:\\
{\renewcommand\baselinestretch{0.5}\selectfont
\begin{eqnarray}
i_t = \sigma(W_i{\boldmath\cdot}[h_{t-1}, x_t]+b_i) & {\quad} & o_t = \sigma(W_o{\boldmath\cdot}[h_{t-1}, x_t]+b_o) 
\end{eqnarray}
\par}
{\renewcommand\baselinestretch{0.1}\selectfont
\begin{eqnarray}
 f_t = \sigma(W_f{\boldmath\cdot}[h_{t-1}, x_t]+b_f) & {\quad} & h_t = o_t * \tanh (s_t)	
\end{eqnarray}
\par}
{\renewcommand\baselinestretch{0.5}\selectfont
\begin{equation}
s_t = f_t*s_{t-1}+i_t*\sigma(W_c{\boldmath\cdot}[h_{c-1}, x_t]+b_c) 
\end{equation}
\par}
\noindent $x_t$ is the current input vector. $h_t$ is the output of cells. $s_t$ is the cell state. $W$, $h$ and $b$ are circulation weight, vector of the current of the current hidden layer and bias.
During inputting tensor of data to the network, we expanded images data from two-dimensional matrixes to one-dimensional vectors to simple the task with ignoring the data’s spatial information. Major steps of the model are shown in Fig.~\ref{fig2}.
\subsection{Band intervals with dense distribution of high identification accuracy determining based on CNN}
\begin{figure}
	\includegraphics[width=\textwidth]{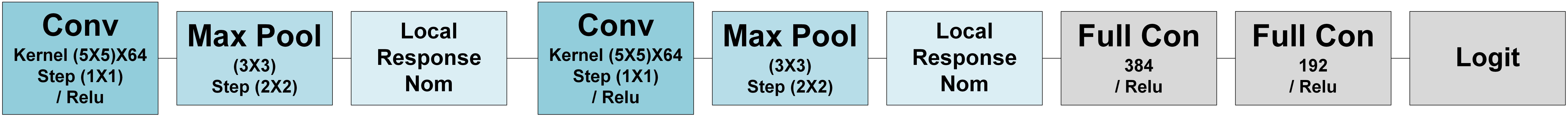}
	\centering
	\caption{The structure of CNN consisting of two typical layers and three full connection layers.} \label{fig3}
\end{figure}

\begin{figure}
	\includegraphics[width=10cm]{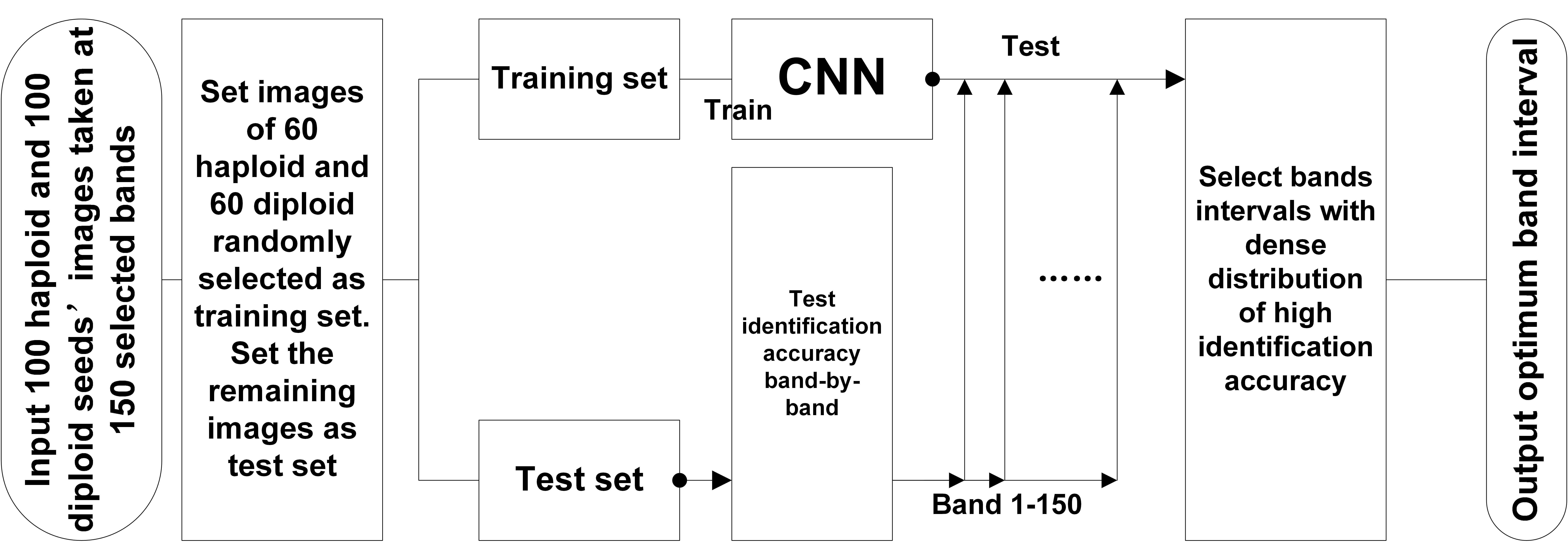}
	\centering
	\caption{Main steps of determining band intervals with dense distribution of high identification accuracy. The accuracy gained through testing test set band-by-band with the trained CNN shows the distribution of the identification accuracy, which helps to determine intervals aggregating optimum hyperspectral images data.} \label{fig4}
\end{figure}

After LSTM removing high-noise hyperspectral data, images of all seed samples in the rest three waveband intervals (about 150 wavebands) include acceptable noises. Although hyperspectral imaging technology can reflect characteristics of different species of maize seeds accurately, it would be applicable to laboratory identification rather than large-scaled seed identification due to the high cost and low scanning speed. A maize haploid seeds identification model based on CNN was constructed in order to lower the cost and simplify the process. This model can determine the waveband interval with dense distribution of high identification accuracy, thus enabling to decrease the number of training samples and optimize quality of training sample data.\\
CNN, as an algorithm based on gradient, which is closer to human vision than LSTM, optimizes the processes of training two-dimensional images and extracts features effectively and abstractly. The features, whose dimension is determined by the size of the convolution kernel and operation of the maximum pooling layer, and which simples ones in shallow layers transfer to complicated ones in deep layers, are transmitted to the softmax to classify haploid and diploid.\\
One typical layer of CNN includes three levels: convolutional layer, detection layer and pooling layer. The convolutional layer combining the detection layer operates convolution to the data ($I$) from the multidimensional array of the previous layer and the kernel function ($K$) and outputs the layer's feature mapping ($S$) processed with linear or nonlinear activation function:$S(i,j)=(I*K)(i,j)$. 
In the pooling layer, the pooling function replaces output of the network at one position by the overall statistical feature of the adjacent output, due to adding one infinitely strong prior and increasing the statistical efficiency of the network significantly. The typical layer has to connect some full connection layers to calculate scores of each type according to previously extracted features. The CNN designed, whose structure is shown in Fig.~\ref{fig3}, consists of two typical layers and three full connection layers. Steps to determine the waveband interval with dense distribution of high identification accuracy are shown in Fig.~\ref{fig4}.
\subsection{Haploid seeds identification and verification}
\begin{figure}
	\includegraphics[width=\textwidth]{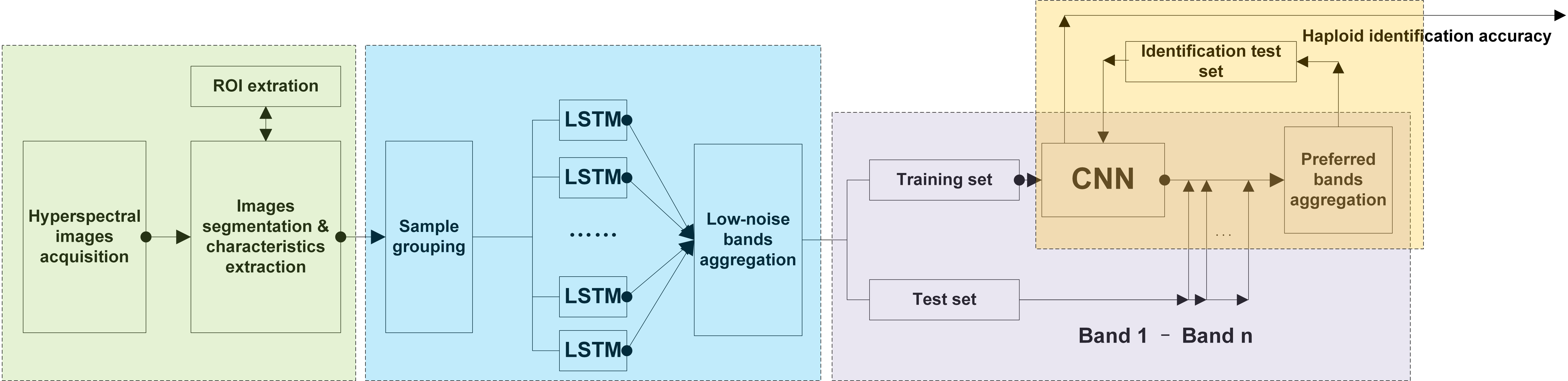}
	\centering
	\caption{The structure that identifies haploid maize seeds with high accuracy and determines bands including optimum hyperspectral images.} \label{fig5}
\end{figure}

Experiments repeat with optimum intervals and determine wavebands with stable high identification accuracy. Images of 60 haploid and 60 diploid seeds selected randomly are set as training set. Images of the rest seeds on the optimum wavebands are used as the test set. The reinitialized CNN is trained by the training set and the final identification accuracy is gained by testing the test set.\\
The verification experiment applies hyperspectral images of 100 haploid seeds and 100 diploid seeds of Nongda-616 on the same optimum waveband intervals, observing whether the law on the other maize cultivar could be similar. The overall model framework is shown in Fig.~\ref{fig5}.

\section{Results}

\subsection{High-noise bands removing based on LSTM}
\begin{figure}
	\includegraphics[width=10cm]{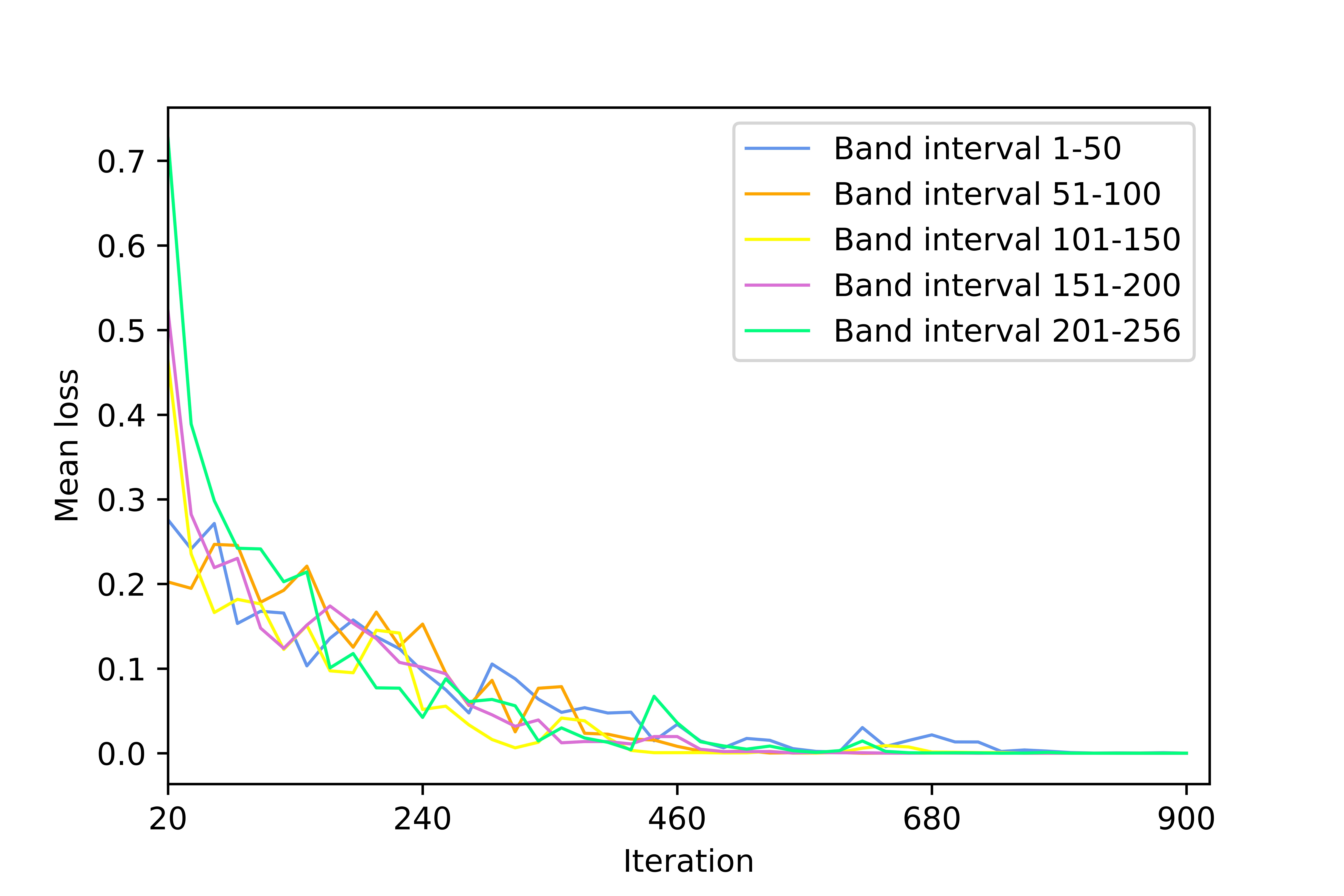}
	\centering
	\caption{The convergence of LSTM's cost functions on band intervals. The iterations of the interval 1-50 (862.9-1031.1nm) and 201-256 (1529.6-1704.2nm) are bigger than the others’. In other word, the two intervals with low convergence rates include more obvious noise than the other intervals.} \label{fig6}
\end{figure}

According to the model requirements, sample images of Zhengdan-958 were divided into five groups among waveband intervals: 1-50 (862.9-1031.1nm), 51-100 (1034.3-1199.2nm), 101-150 (1202.5-1364.4nm), 151-200 (1367.6-1526.4nm) and 201-256 (1529.6-1704.2nm). In actual experiment, the hidden units, the size of batch, time step and learning rate for Adam were set 128, 128, 128 and 0.001, respectively. The total number of training iterations was 5,000. After three repeated experiments in each waveband interval, the curves of mean loss values under different numbers of iterations were drawn (Fig.~\ref{fig6}), based on which the number of iterations at convergence of cost functions convergence on five waveband intervals are 760, 540, 460, 580 and 660, respectively. The convergence rate of the cost function corresponding to data at two ends of the test waveband intervals is relatively low, showing these bands' data with distinct noises. This finding conforms to the experimental observation that there are great noises at two ends of the response interval of the spectrometer sensor. Therefore, image data of the Zhengdan-958 seeds taken at the waveband interval of 1-50 and 201-256 were removed from the follow-up experiment.
\subsection{Band intervals with dense distribution of high identification accuracy determining model based on CNN}
\begin{figure}[!ht] 
	\begin{center}
		\includegraphics[width=6cm,keepaspectratio,]{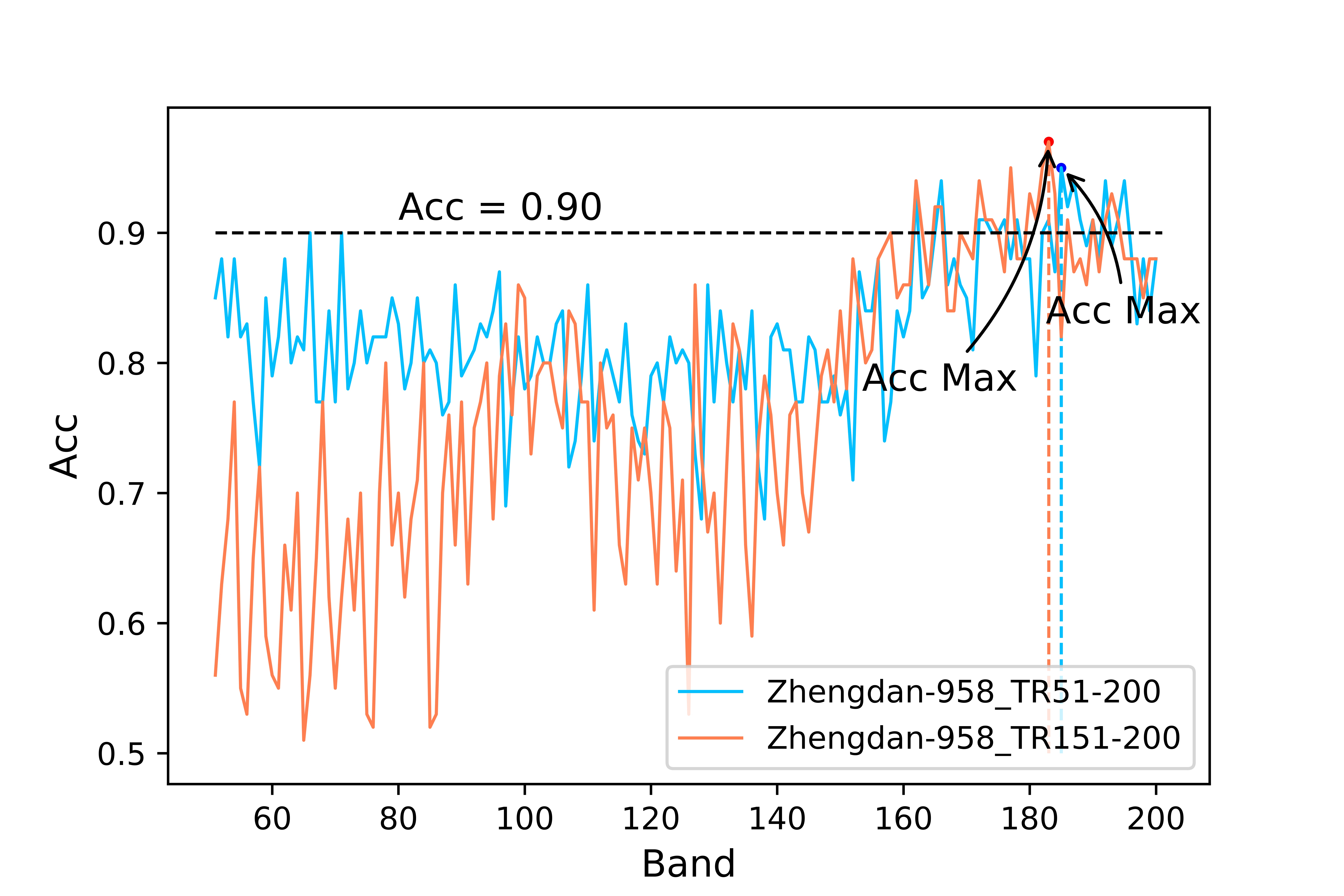}
		\includegraphics[width=6cm,keepaspectratio,]{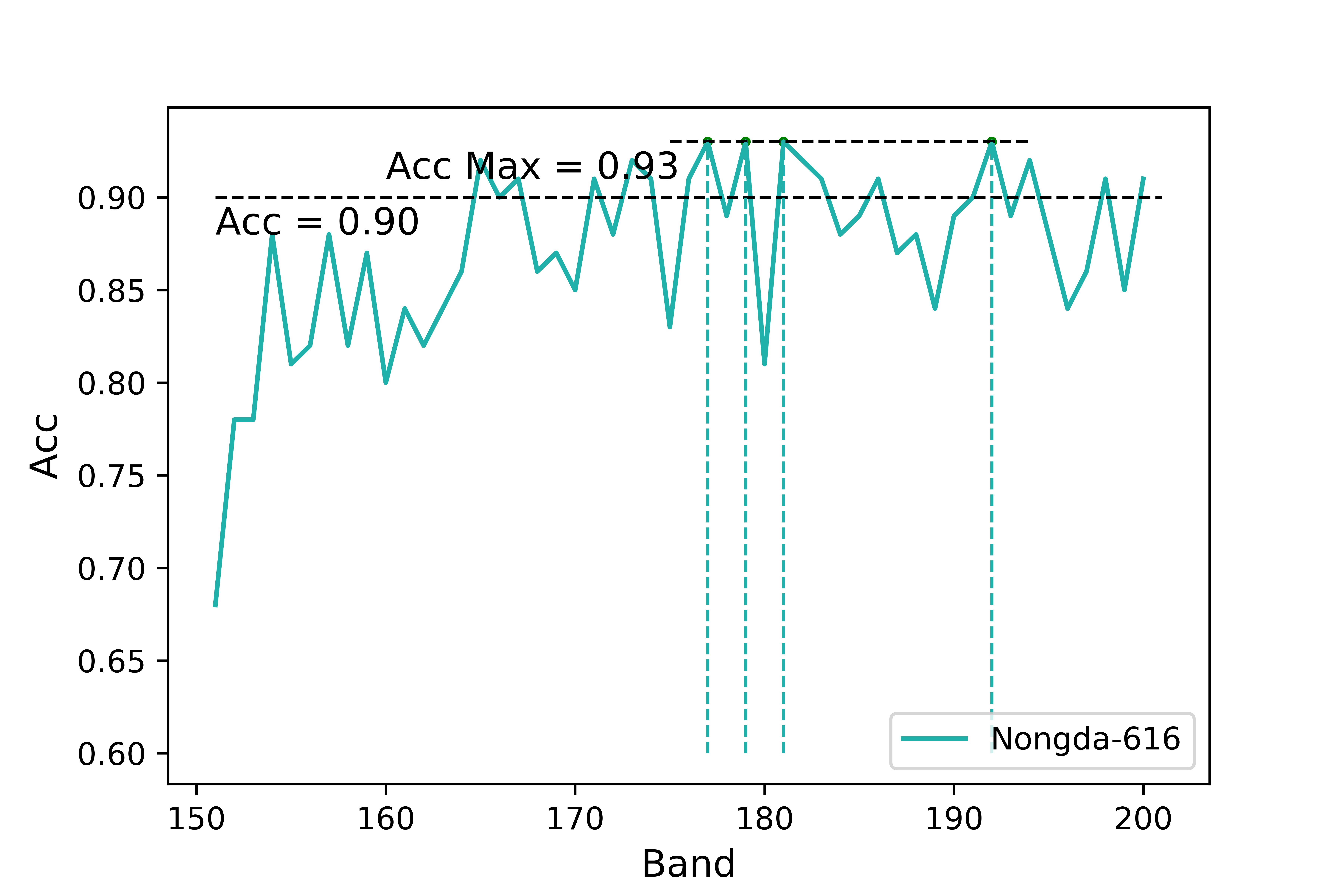}
		\caption{(Left.)Test set’s identification accuracy gained through CNN, which trained by training set of interval 51-200 (1034.3-1526.4nm) and 151-200 (1367.6-1526.4nm). The testing accuracy fluctuation relating to the model trained by 51-200 interval data highlights the interval of 151-200 where optimum images data aggregate. These preferred images encourage the identification accuracy and stability of the interval 151-200 improving when being applied to training the CNN, though these ones don’t improve the testing behaviors on the other intervals that are unhelpful to identifying haploid seeds. } \label{fig7}
		\caption{(Right.)Verification. Experimental verification used Nongda-616's hyperspectral images taken at 151-200 interval behaves the similar identification, which proves the model's generalization.} \label{fig8}
	\end{center}
\end{figure}

Hyperspectral images of 100 haploid and 100 diploid seeds of Zhengdan-958 on the waveband intervals of 51-200 were divided into the training set and test set according to model requirements. The size of batch and learning rate for Adam were set 128 and 0.001, respectively. The total number of training iterations steps was set 5,000. The identification accuracy of the test set on each waveband is expressed by the blue curve in Fig.~\ref{fig7}. The mean, maximum, standard deviation and the number of wavebands' identification accuracy ($\geq$0.90) in three waveband intervals (51-100, 101-150 and 151-200) are shown in Table~\ref{tab2}. Obviously, the highest identification accuracy (0.95) appeared on the waveband interval of 151-200, whose mean and waveband number of identification accuracy ($\geq$0.90) also were the highest. Therefore, the waveband interval of 151-200 was the band intervals with dense distribution of high identification accuracy.
\begin{table}
	\centering
	\caption{The statistical index of testing CNN model.}\label{tab2}
	\begin{tabular}{c c c c c}
		\toprule
		{Band interval} & {Mean} & {Max} & {Standard deviation} & {Num of acc $\geq$ 0.90}\\
		\midrule
		51-100 & 0.814 & 0.90 & 0.0419 & 2\\
		101-150 & 0.787 & 0.86 & 0.0409 & 0\\
		151-200 & 0.872 & 0.95 & 0.0514 & 18\\
		\bottomrule 
	\end{tabular}
\end{table}
\subsection{Haploid seeds identification}
 Images on the waveband interval of 151-200 were divided into the training set and the test set as the model requirements. Identification accuracy of the test set was tested after the reinitialized network being trained. Results are shown in red curve in Fig.7. The mean, maximum, standard deviation and the number of wavebands' identification accuracy ($\geq$ 0.90) of three waveband intervals of 51-100, 101-150 and 151-200 are shown in Table~\ref{tab3}. Comparison between Table~\ref{tab2} and Table~\ref{tab3} argues that the identification accuracy increases significantly when hyperspectral images taken at wavebands of 151-200 instead of all data at bands of 51-200 trained CNN. The highest identification accuracy and the number of bands with accuracy $\geq$0.90 incresed to 0.97 and 21, respectively. Moreover, the standard deviation declines dramatically, indicating the better stability of identification. Based on multiple repeated experiments, the wavebands with identification accuracy $\geq$0.90 are 165 (1413.3nm), 166 (1416.6nm), 173 (1439.3nm), 174 (1442.5nm), 182 (1468.4nm), 183 (1471.7nm), 186 (1481.4nm) and 192 (1500.7nm). The final identification accuracy of haploid seeds was 0.97 when modeling with hyperspectral data of abovementioned wavebands.
 
 \begin{table}
 	\centering
 	\caption{The statistical index of 151-200 testing haploid seeds identification model.}\label{tab3}
 	\begin{tabular}{c c c c c}
 		\toprule
 		{Band interval} & {Mean} & {Max} & {Standard deviation} & {Num of acc $\geq$ 0.90}\\
 		\midrule
 		51-100 & 0.6682 & 0.86 & 0.0974 & 0\\
 		101-150 & 0.7332 & 0.86 & 0.0727 & 0\\
 		151-200 & 0.8858 & 0.97 & 0.0393 & 21\\
 		\bottomrule
 	\end{tabular}
 \end{table}

\subsection{Verification}
To verify the universality, experimental verification applied hyperspectral images of Nongda-616 on the waveband interval of 151-200. Identification accuracy of the test set was tested after the reinitialized network being trained. Results are shown in Fig.~\ref{fig8}. The mean, maximum, standard deviation and the number of wavebands' identification accuracy ($\geq$0.90) are 0.8704, 0.93, 0.049361639 and 18, respectively. It proves that interval of 151-200 certainly aggregates optimum data through comparing with Table 3. Correlation coefficient of two identification accuracy vectors of Zhengdan-958 and Nongda-616 on the waveband interval of 151-200 calculated at 0.433875 indicates the consistence of variations between two cultivars' identification. Wavebands with identification accuracy $\geq$0.90 were combined for modeling test, achieving an identification accuracy of 0.93. Verification succeeded.\\
These results demonstrated that the proposed model not only can identify maize haploid seeds accurately, but also can determine the optimum wavebands of hyperspectral images information, which benefits other similar sorting experimentations via hyperspectral imaging from collecting data with references, thus reducing cost obviously.

\section{Discussion}
With respect to maize haploid seed identification based on the hyperspectral imaging technology, the LSTM-CNN determined the optimum wavebands data, and encouraged training the model to achieve high identification accuracy. The training process of LSTM is an iterative optimization based on gradient and is due to making the cost function, expressing the cross entropy between the training data and model prediction in regularization neural networks, converges to the minimum value. The convergence speed is related with the complexity of probability distribution of sample data. Therefore, in five waveband intervals, the wavebands with lower convergence rate included more distinct noise interferences, which have to be eliminated. CNN determined the waveband intervals with dense distributions of high identification accuracy through testing on the rest wavebands band-by-band. By comparing the Table 2 and Table 3, the neural network trained by images from waveband interval of 151-200 (1367.6-1526.4nm) is better than that trained by ones form the waveband interval of 51-200 (1034.3-1526.4nm) in term of test stability and identification accuracy. The improvement states that seeds information collected from different wavebands is uneven. Network training based on the hyperspectral imaging data with the best seed information acquisition is conducive to benefits learning feature from the identification model. Testing on the other cultivar draw on the same conclusion, which proves the generalization of the proposed identification model.\\
A comparison of our LSTM-CNN model with Yang et al. (2015), Huang et al. (2016) and Cui (2017) claims that noises in hyperspectral images affect the identification accuracy and the supervised strategy works on removing the interferences. That is to say, complicated noises and uneven information of the hyperspectral imaging technology challenge the sorting. For a vivid interpretation, the grayscale histograms of hyperspectral images of 4 randomly selected maize seeds taken at wavebands of 30 (962.7nm), 80 (1132.3nm), 130 (1298.7nm), 180 (1462.0nm) and 230 (1622.1nm) are shown in Fig.~\ref{fig9}. Data collected by the hyperspectral imaging technology are high-dimensional tensor with complicated noises (Liu et al. 2012). SVM and its variant algorithms are widely used, because they can process such high-dimensional nonlinear problems in some way, where SVM views the decision function as the line function in different spaces by applying the kernel technique, thus enabling to learn the nonlinear infinite-dimensional model by using the convex optimization function that can ensure effective convergence (Boser et al. 1992; Cortes et al. 1995). Nevertheless, the mapping between the linear function and nonlinear space has to be designed manually, resulting in the poor generalization. The computational load of the kernel machine increases with the growth of dataset. These bottlenecks challenge expanding applications of SVM. The linear layer function of the neural networks treats the mapping of the input data ($\O(x)$) as the operation object, where $\O$ is the universal nonlinear variable mapping hidden in the kernel machine and enhances the generalization of the neural network (Goodfellow et al. 2016), which is different from the linear kernel technique and is the main reason of the generalization of neural network-based identification model.\\
Essentially, these methods all belong to supervised algorithms. If the input and output of the dataset were given, it has to design labels or indexes to assist the model learning. On the contrary, the unsupervised algorithms just process features without extracting information from artificial annotated samples. To establish a universal hyperspectral classification model, the unsupervised algorithms shall be introduced into the model. Gaining features by unsupervised algorithms can optimize the CNN. Coates et al. (2011) has applied the k-means clustering algorithm on small-sized images and used every extracted center as the convolutional kernel, achieving satisfying effect. So it is feasible to customize the convolutional kernel through unsupervised learning, which is also beneficial to establish a complicated deep network without generating the vanishing gradient. 
{\renewcommand\baselinestretch{0.5}\selectfont
\begin{figure}
	\includegraphics[width=8cm]{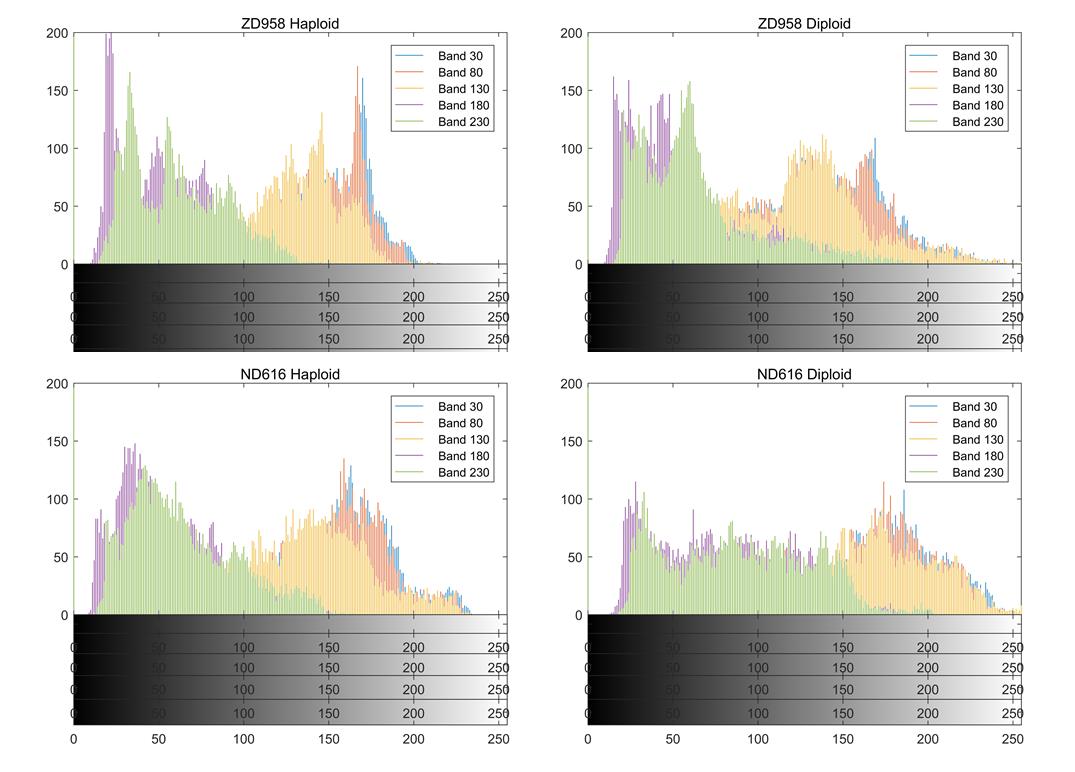}
	\centering
	\caption{The grayscale histograms of seeds' hyperspectral images. The same seed's images taken at different wavebands show variable distributions of gayscale, since the images are made different from each other by the uneven image information and noise.} \label{fig9}
\end{figure}
\par}
\section{Conculsion}
 In this study, deep learning and hyperspectral imaging technology are used to identify the artificially induced maize haploid seeds from abundant maize diploid seeds with high accuracy. The hyperspectral imaging system collects feature information of 256 wavebands of maize haploid and diploid seeds in the range of 862.9-1704.2nm. LSTM removed the waveband intervals including obviously high noises by contrasting different waveband intervals’ iterations at convergence of the cost function with each other during training network. CNN determined the waveband intervals with dense distribution of high identification accuracy by testing the rest wavebands' identification accuracy band-by-band. The identification accuracy of the proposed model on the optimum test set reaches 97\%, which is higher than the accuracy of modeling based on any one of the other waveband intervals ($<$90\%) or based on the full waveband interval (80\%). The model can meet demands of actual sorting with high accuracy and determine waveband interval of 1367.6-1526.4nm containing the optimum sample data, which benefits collecting data at the interval first instead of collecting samples at all bands expensively and time-consuming. All bands data even maybe weaken the identification.

\end{document}